\pdfoutput=1

\documentclass[11pt]{article}

\usepackage{acl}

\usepackage{times}
\usepackage{latexsym}

\usepackage[T1]{fontenc}

\usepackage[utf8]{inputenc}

\usepackage{microtype}

\usepackage{hyperref}       
\usepackage{url}            
\usepackage{booktabs}       
\usepackage{amsfonts}       
\usepackage{nicefrac}       
\usepackage{graphicx}
\usepackage{amsmath}
\usepackage{amsfonts}
\usepackage{amssymb}
\usepackage{amsthm}
\usepackage{amsmath,bm}
\usepackage{natbib}
\usepackage{subfigure,graphicx}
\usepackage{paralist}
\usepackage{mdwlist}
\usepackage{wrapfig}
\usepackage{xcolor}
\usepackage{multirow}
\usepackage{booktabs}
\usepackage{makecell}
\usepackage{xspace}
\usepackage{caption}
\usepackage{textcomp}
\usepackage{stfloats}

\newcommand{\paratitle}[1]{\noindent\textbf{#1}}

\newcommand{\method}{TACO\xspace}
\newcommand{\objective}{TC\xspace}

\title{Contextual Representation Learning beyond Masked Language Modeling}

\author{

    Zhiyi Fu\textsuperscript{1}\thanks{\ \ Equal Contribution} \ \thanks{\ \ This work is done at ByteDance AI Lab.}, 
    \ Wangchunshu Zhou\textsuperscript{2}\footnotemark[1],
    \ Jingjing Xu\textsuperscript{3}\footnotemark[2], 
    \ Hao Zhou\textsuperscript{2}, 
    \ Lei Li\textsuperscript{3}\footnotemark[2]\\
    
    \textsuperscript{1}Peking University\ \ 
    \textsuperscript{2}ByteDance AI Lab\ \ 
    \textsuperscript{3}University of California, Santa Barbara \\
    
    \texttt{ypfzy@pku.edu.cn} \\
    \texttt{\{zhouwangchunshu.7, zhouhao.nlp\}@bytedance.com} \\
    \texttt{\{jingjingxu, leili\}@cs.ucsb.edu} \\
}

\begin{document}
\maketitle

\begin{abstract}

How do masked language models (MLMs) such as BERT learn contextual representations?
In this work, we analyze the learning dynamics of MLMs. 
We find that MLMs adopt sampled embeddings as anchors to estimate and inject contextual semantics to representations, which limits the efficiency and effectiveness of MLMs. 
To address these issues, we propose \method, a simple yet effective representation learning approach to directly model global semantics. 
\method extracts and aligns contextual semantics hidden in contextualized representations to encourage models to attend global semantics when generating contextualized representations.
Experiments on the GLUE benchmark show that \method achieves up to 5x speedup and up to 1.2 points average improvement over existing MLMs. The code is available at \url{https://github.com/FUZHIYI/TACO}.
\end{abstract}

\section{Introduction}
\label{sec:intro}

In the age of deep learning, the basis of representation learning is to learn distributional semantics. The target of distributional semantics can be summed up in the so-called distributional hypothesis~\citep{harris1954distributional}: \textit{Linguistic items with similar distributions have similar meanings}. 
To model similar meanings, traditional representation approaches~\citep{DBLP:conf/nips/MikolovSCCD13,DBLP:conf/emnlp/PenningtonSM14} (e.g., Word2Vec) model  distributional semantics by defining tokens using \textit{context-independent} (CI) dense vectors, i.e., word embeddings, and directly aligning the representations of tokens in the same context. 
Nowadays, pre-trained language models (PTMs)~\citep{bert,radford2018improving,DBLP:journals/corr/abs-2003-08271} expand static embeddings into contextualized representations where each token has two kinds of representations: \textit{context-independent} embedding, and \textit{context-dependent} (CD) dense representation that stems from its embedding and contains context information. Although  language modeling and representation learning have distinct targets,  masked language modeling is still the prime choice to learn token representations with access to large scale of raw texts~\citep{DBLP:conf/naacl/PetersNIGCLZ18,bert,DBLP:journals/jmlr/RaffelSRLNMZLL20,DBLP:conf/nips/BrownMRSKDNSSAA20}. 

\begin{figure}[t]
\centering
\small
     \includegraphics[width=1\linewidth]{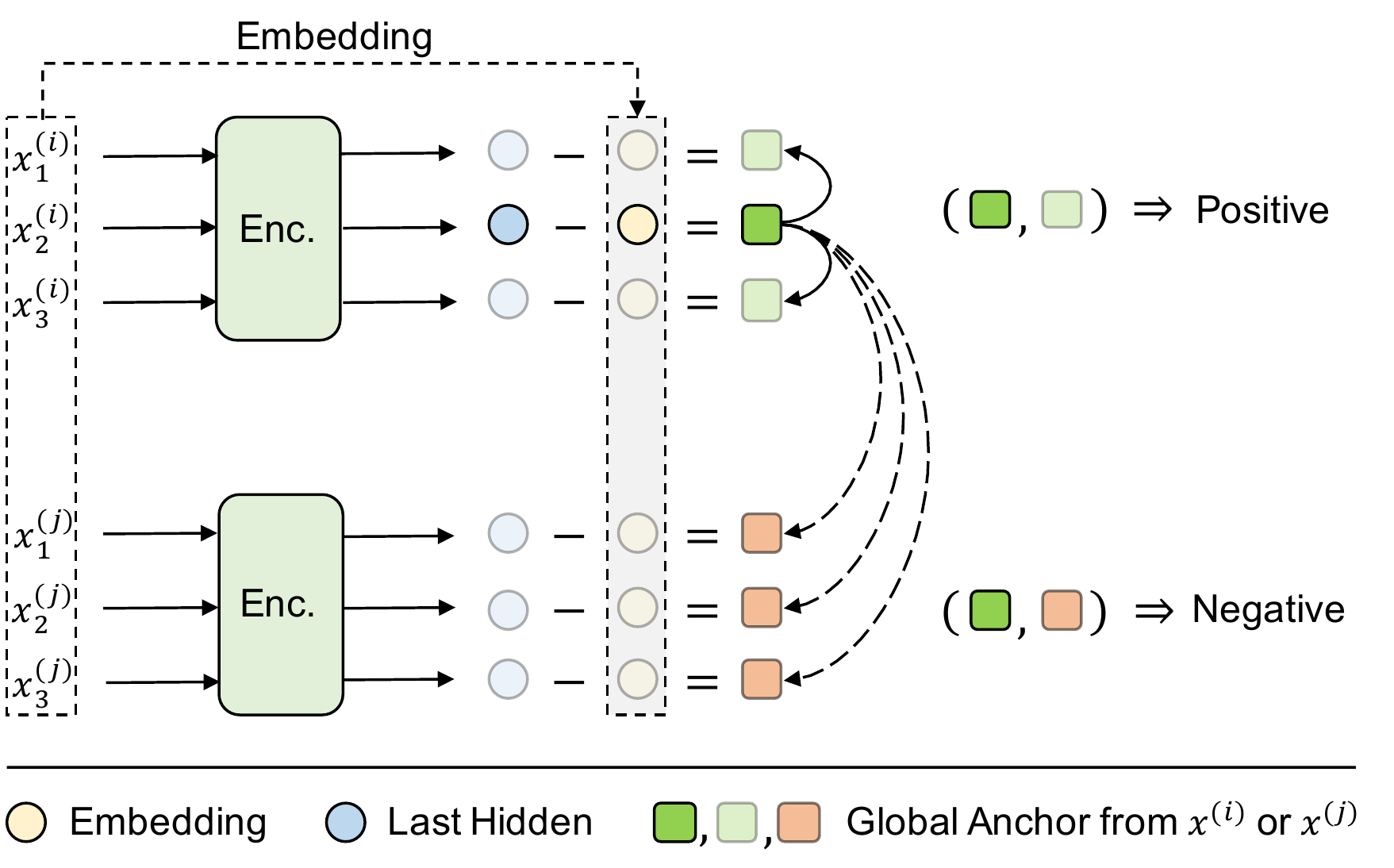}
     \caption{Illustration of the proposed token-alignment contrastive objective. It extracts and aligns the global semantics hidden in contextualized representations via the gap between contextualized representations and corresponding static embeddings.}
 \label{fig:framework}
 \end{figure}

It naturally raises a question: How do masked language models learn contextual representations? 
Following the widely-accepted understanding~\citep{DBLP:conf/icml/0001I20}, MLM optimizes two properties, the alignment of contextualized representations with the static embeddings of masked tokens, and the uniformity of static embeddings in the representation space. In the alignment property, sampled embeddings of masked tokens play as an \textit{anchor} to align contextualized representations. We find that although such local anchor is essential to model local dependencies, the lack of global anchors brings several limitations. First, experiments show that the learning of contextual representations is sensitive to embedding quality, which harms the efficiency of MLM at the early stage of training. Second, MLM typically masks multiple target words in a sentence, resulting in multiple embedding anchors in the same context. This pushes contextualized representations into different clusters and thus harms modeling global dependencies. 

To address these challenges, we propose a novel \textbf{T}oken-\textbf{A}lignment \textbf{C}ontrastive \textbf{O}bjective (\textbf{TACO}) to directly build global anchors. By combing local anchors and global anchors together, TACO achieves better performance and faster convergence than MLM. Motivated by the widely-accepted belief that contextualized representation of a token should be the mapping  of its static embedding on the contextual space given  global information, we propose to directly align global information hidden in contextualized representations at all positions of a natural sentence to encourage models to attend same global semantics when generating contextualized representations.  Concerning possible relationships between context-dependent and context-independent representations, we adopt the simplest probing method to extract global information via the gap between context-dependent and context-independent representations of a token for simplification, as shown in Figure~\ref{fig:framework}. To be specific, we define tokens in the same context (text span) as positive pairs and tokens in different contexts as negative pairs, to encourage the global information among tokens within the same context to be more similar compared to that from different contexts.

We evaluate TACO on GLUE benchmark.  Experiment results show that \method outperforms MLM with average 1.2 point improvement and 5x speedup (in terms of sample efficiency) on BERT-small, and with average 0.9 point improvement and 2x speedup on BERT-base. 
 
The contributions of this paper are as follows.
\begin{itemize}
    \item We analyze the limitation of MLM and propose a simple yet efficient method \method to directly model global semantics. 
    \item Experiments show that \method outperforms MLM with up to 1.2 point improvement and up to 5x speedup on GLUE benchmark. 
\end{itemize}

\section{Understanding Language Modeling}
\label{sec:semgrav}

\subsection{Objective Analysis}

The key idea of MLM is to randomly replace a few tokens in a sentence with the special token \texttt{[MASK]} and ask a neural network to recover the original tokens. 
Formally, we define a corrupted sentence as $\bm{x}_1$, $\bm{x}_2$, $\cdots$, $\bm{x}_L$, and feed it into a Transformers encoder~\citep{DBLP:conf/nips/VaswaniSPUJGKP17}, the hidden states from the final layer are denoted as $\bm{h}_1$, $\bm{h}_2$, $\cdots$, $\bm{h}_L$. We denote the embeddings of the corresponding original tokens as $\bm{e}_1$, $\bm{e}_2$, $\cdots$, $\bm{e}_L$. The MLM objective can be formulated as:
\begin{equation}\label{mlmloss1}
\footnotesize
\mathcal{L}_{\text{MLM}}(\bm{x})=-\frac1{|\mathcal{M}|}\sum_{i\in\mathcal{M}}\log\frac
{\exp(\bm{m}_i \cdot \bm{e}_i)}
{ \sum_{k=1}^{|\mathcal{V}|} \exp(\bm{m}_i \cdot \bm{e}_k)}
\end{equation}
where $\mathcal{M}$ denotes the set of masked tokens and $|\mathcal{V}|$ is the size of vocabulary $\mathcal{V}$. $\bm{m}_i$ is hidden state of the last layer at the masked position, and can be regarded as a fusion of contextualized representations of surrounding tokens. 
Following the widely-accepted understanding~\citep{DBLP:conf/icml/0001I20}, Eq.1 optimizes: (1) the alignment between contextualized representations of surrounding tokens and the context-independent embedding of the target token and (2) the uniformity of representations in the representation space.

\begin{figure}[!ht]
\centering
\small
     \includegraphics[width=1\linewidth]{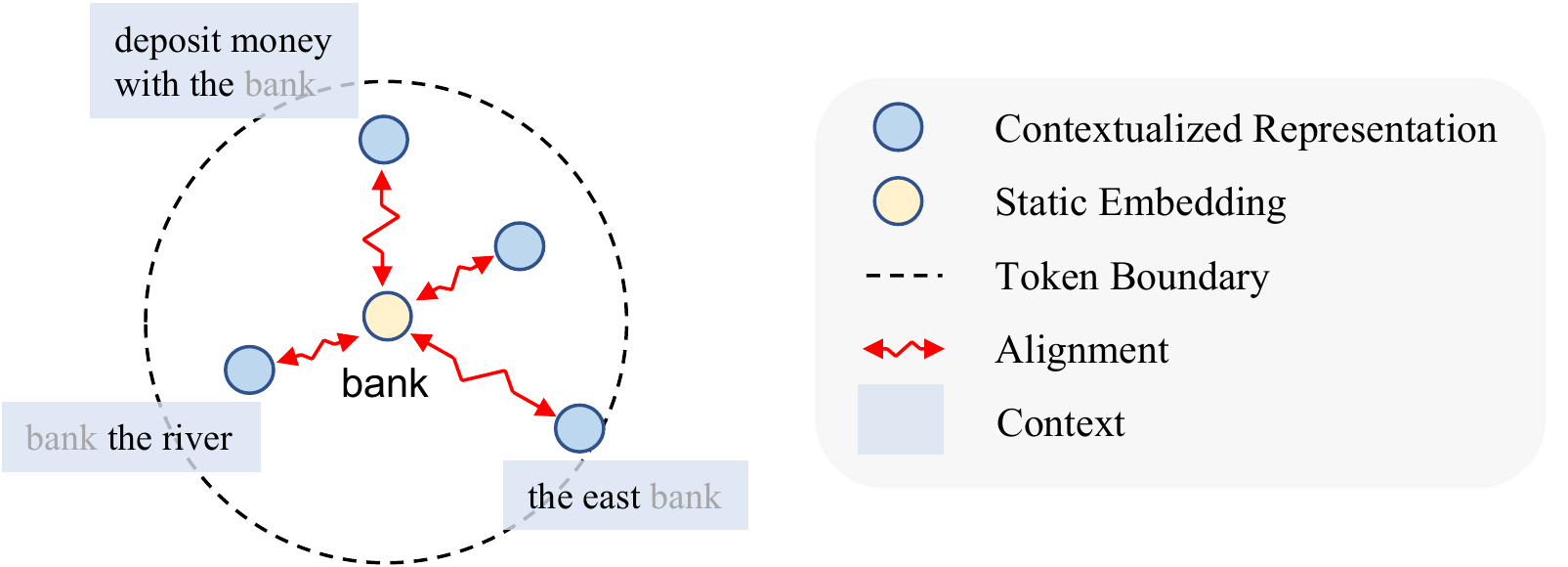}
     \caption{Illustration of the MLM objective. At the alignment part, it uses static embedding of masked tokens to align contextualized representations in the same context. }
 \label{fig:MLM}
 \end{figure}

In the alignment part, MLM relies on sampled contextual-independent embeddings of masked tokens as anchors to align contextualized representations in contexts, as shown in Figure~\ref{fig:MLM}.  Local anchor is the key feature of MLM. Therefore, the learning of contextualized representations heavily relies on embedding quality. In addition, multiple local anchors in a sentence tend to pushing contextualized representations of surrounding tokens closer to different clusters, encouraging models to attend local dependencies where global semantics are neglected. 

\subsection{Empirical Analysis}
\label{sec:analysis}

To verify our understanding, we conduct comprehensive experiments to investigate: How does embedding anchor affect the learning dynamics of MLM? We re-train a BERT-small~\citep{bert} model with the MLM objective solely and analyze the changes in its semantic space during pre-training. The training details are described in Appendix A.

\begin{figure}[tb]
\centering
\small
\begin{minipage}[b]{\linewidth}
    \centering
    \includegraphics[width=0.7\linewidth]{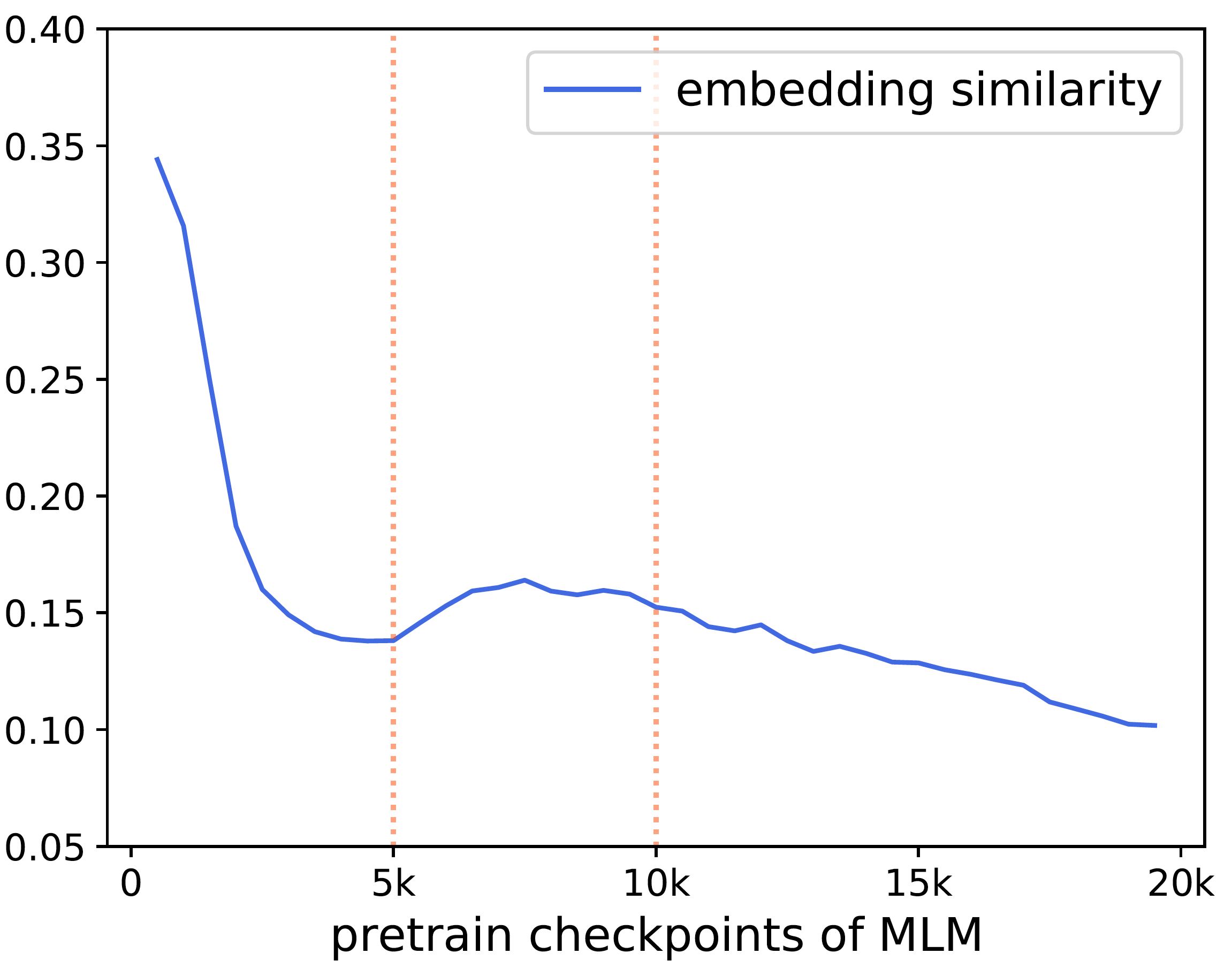}
    \centering
     \includegraphics[width=0.7\linewidth]{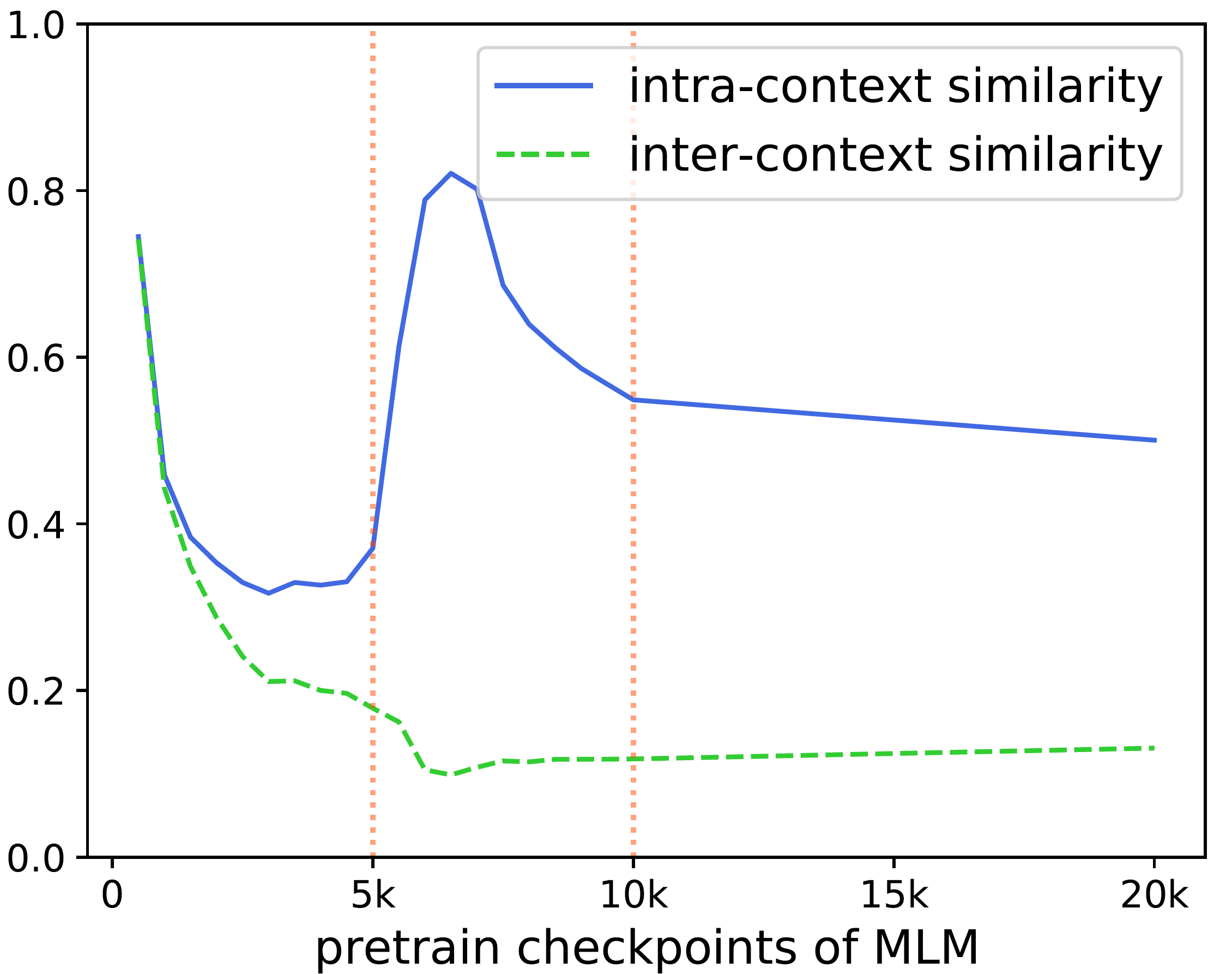}
   
 \end{minipage}

\caption{The learning dynamics of MLM. The top figure (a) illustrates the similarity between embeddings of frequently co-occurrent tokens (e.g., bank and money). The bottom figure (b) illustrates the similarity between contextualized representation of tokens from the same context and different contexts. These figures show an embedding bias problem where only the randomly selected target embeddings in MLM are similar, contextualized representations in the same context will be aligned with similar features.
}

\label{fig:similarity}
\end{figure}

\paragraph{Contextualized representation evaluation.} In general, if contextualized representations are well learned, the contextualized representations in a same context will have higher similarity than that of in different contexts. Naturally, we use the gap between intra-sentence similarity and inter-sentence similarity to evaluate contextual information in contextualized representations. We call this gap as \textit{contextual score}. The similarity can be evaluated via probing methods like L2 distance, cosine similarity, etc. We observe similar findings on different probing methods and only report cosine similarity here for simplification.
Figure~\ref{fig:similarity}(b) shows how contextual score changes during training. Other statistical results are listed in Appendix A.

\paragraph{Embedding similarity evaluation.} To observe how sampled embeddings affect contextualized representation learning, we evaluate the embedding similarity between co-occurrent tokens. Motivated by the target that co-occurrent tokens should have similar representations, we use the similarity score calculated by cosine similarity between co-occurrent words labeled by humans (sampled from the WordSim353 dataset~\citep{agirre2009study}) as the evaluation metric. Figure~\ref{fig:similarity}(a) shows how embedding similarity between co-occurrent tokens changes during training.

\paratitle{The learning of contextualized representations heavily relies on embeddings similarity. }  As we can see from Figure~\ref{fig:similarity}(a), the embedding similarity between co-occurrent tokens first decreases during the earliest stage of pre-training. It is because all embeddings are randomly initialized with the same distribution and the uniformity feature in MLM pushes tokens far away from each other, thus resulting in the decrease of embedding similarity. Meanwhile, the contextual score, i.e., the gap between intra-context similarity and inter-context similarity in Figure~\ref{fig:similarity}(b), does not increase at the earliest stage of training. It shows that random embeddings provide little help to learn contextual semantics.  During 5K-10K iterations, only when embeddings become closer, contextualized representations in the same context begin to have similar features. At this stage, the randomly sampled embeddings from the same sentence, i.e., the same context, usually have similar representations and thus MLM can push contextualized tokens closer to each other.

\begin{figure}[h]
\centering
\small
\subfigure{
    \begin{minipage}[b]{0.7\linewidth}
    \includegraphics[width=1\linewidth]{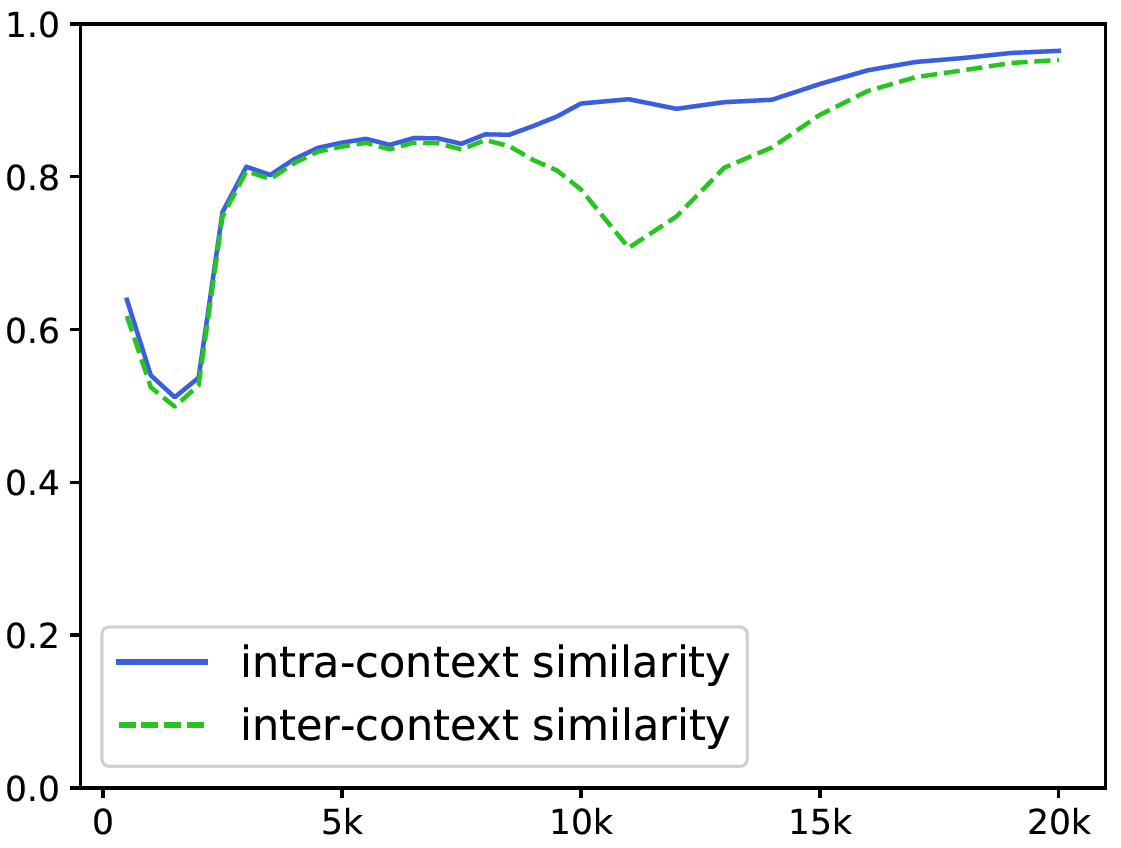}
    \end{minipage}
}\hspace{8pt}
\subfigure{
    \begin{minipage}[b]{0.7\linewidth}
    \includegraphics[width=1\linewidth]{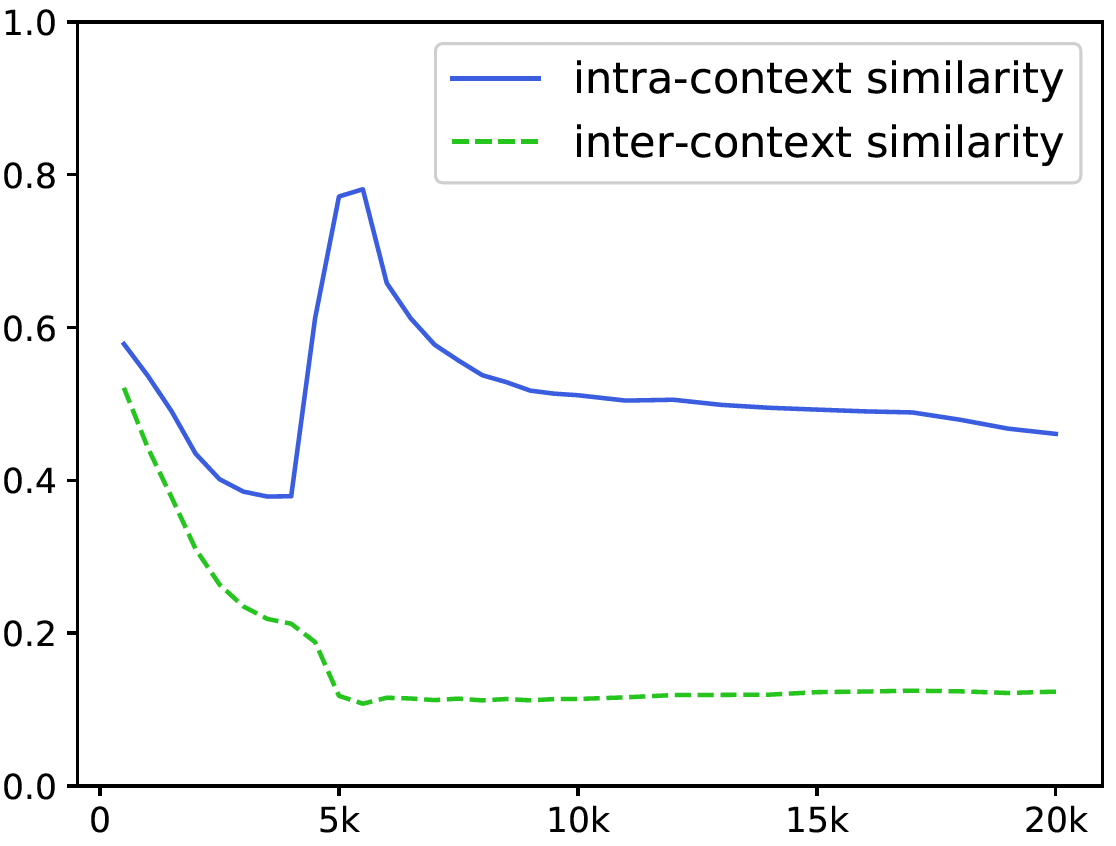}
    \end{minipage}
}

\caption{The impact of embedding quality for the learning of contextualized representations. We train two BERT-small variants from scratch, whose embedding is either (a) randomly initialized and frozen or (b) copied from normally pre-trained BERT at 250k steps and frozen.}

\label{fig:embedding_curse}
\end{figure}

We further verify the effects of embedding quality in Figure \ref{fig:embedding_curse}. To this end, we train two BERT models whose embedding matrices are frozen and initialized with the ones from different pre-training stage.  We can see the model initialized with random embedding fails to teach contextualized representations to attend sentence meanings and representations from different contexts have almost the same similarity. However, the variant with well-trained but frozen embeddings learns to distinguish different contexts early at around 4k steps. These statistical observations verify that embedding anchors bring the efficiency and effectiveness problem. 
 
\paragraph{Surprisingly, embedding anchors reduce global contextual information in contextualized representation at the later stage of training. }  
Figure~\ref{fig:similarity}(a) shows that embedding similarity begins to drop after  8k steps. It shows that the model learns the specific meanings of co-occurrent tokens and begins to push them a little bit far away. Since MLM adopts local anchors, these local embeddings push contextualized representations into different clusters. The contextual score begins to decrease too.  This phenomenon proves the embedding bias problem where the learning of contextualized representations is decided by the selected embeddings where the global contextual semantics are neglected.

\section{Proposed Approach: \method}
\label{sec:approach}

To address the challenges of MLM, we propose a new method \method to combine global anchors and local anchors. We first introduce \objective, a token-alignment contrastive loss which explicitly models global semantics in Section 
\ref{subsec:semgrav:1}, and combine \objective with MLM to get the overall objective for training our \method model in Section \ref{subsec:semgrav:2}.

\subsection{Token-alignment Contrastive Loss}
\label{subsec:semgrav:1}

To model global semantics, the objective is expected to be capable of explicitly capturing information shared between contextualized representation of tokens within the same context.
Therefore, a natural solution is to maximize the mutual information of contextual information hidden in  contextualized representations in the same context. To extract shared contextual information, we first define a rule to generate contextual representations of tokens by combining embeddings and global information. Formally,
\begin{equation}
\small
\bm{h}_i = f(\bm{e}_i, \bm{g}).
\end{equation}
where $f$ is a probing algorithm and $\bm{e}_i$ is the  embedding and $\bm{g}$ is the global bias of a concrete context. In this paper, we adopt a straightforward probing method to get global information hidden in contextualized representations, where
\begin{equation}
\small
\label{eq:bias}
\bm{g}_i = \bm{h}_i - \bm{e}_i.
\end{equation}

Given contextualized representations of an token $\bm{x}$ and its nearby tokens $\bm{c}$ in the same context, we use $\bm{g}_x$ and $\bm{g}_c$ to represent global semantics hidden in these representations. The mutual information between the two global bias $\bm{g}_x$ and $\bm{g}_c$ is
\begin{equation}
\small
    I(\bm{g}_x,\bm{g}_c) = \sum_{\bm{g}_x,\bm{g}_c}p(\bm{g}_x,\bm{g}_c)\log\frac{p(\bm{g}_x|\bm{g}_c)}{p(\bm{g}_x)}
\end{equation}
According to~\citealt{oord2019representation}, the InfoNCE loss serves as an estimator of mutual information of $\bm{x}$ and $\bm{c}$:
\begin{equation}
\small
\label{mi}
     I(\bm{g}_x,\bm{g}_c)  \geq \log(K) - \mathcal{L}(\bm{g}_x,\bm{g}_c)
\end{equation}
where $\mathcal{L}(\bm{g}_x,\bm{g}_c)$ is defined as:
\begin{equation}
\small
  \mathcal{L}(\bm{g}_x,\bm{g}_c) = -\mathbb{E} \left[ \log\frac{f(\bm{g}_x,\bm{g}_c)}{f(\bm{g}_x,\bm{g}_c) + \sum_{k=1}^{K}f(\bm{g}_x, \bm{g}_{c_k^-})} \right] 
\end{equation}
where $\bm{c}_k^-$ is the $k$-th negative sample of $\bm{x}$ and $K$ is the size of negative samples. Hence minimizing the objective $  \mathcal{L}(\bm{g}_x,\bm{g}_c)$ is equivalent to maximizing the lower bound on the mutual information $ I(\bm{g}_x,\bm{g}_c)$. This objective contains two parts: \textit{positive pairs} $f(\bm{g}_x,\bm{g}_c)$ and \textit{negative pairs} $f(\bm{g}_x, \bm{g}_{c_k^-})$. 

Previous study~\citep{DBLP:conf/icml/ChenK0H20} has shown that cosine similarity with temperature performs well as the score function $f$ in InfoNCE loss. Following them, we take
\begin{gather}
\small
f(\bm{g}_x, \bm{g}_c) = \frac1\tau\frac{\bm{g}_x\cdot \bm{g}_c}{\Vert\bm{g}_x\Vert \Vert\bm{g}_c\Vert}
\end{gather}
where $\tau$ is the temperature hyper-parameter and $\Vert\cdot\Vert$ is $\ell_2$-norm function.

\textit{Contextualized representation}: To get global bias $\bm{g}_x$ and $\bm{g}_c$ following Eq.~\ref{eq:bias}, we adopt the widely-used Transformer~\citep{DBLP:conf/nips/VaswaniSPUJGKP17} as the encoder and take the last hidden states as the contextualized representations $\bm{h}_x$ and $\bm{h}_c$. 
Formally, suppose a batch of sequences $\{ \bm{s}_i \}$ where $i\in\{1,\cdots,N\}$.
We feed it into the Transformer encoder to obtain contextualized representations, $\bm{h}_1^i$, $\bm{h}_2^i$, $\cdots$, $\bm{h}_{|\bm{s}_i|}^i$ where $\bm{h}_j^i \in \mathbb{R}^d$.

\textit{Positive pairs}: Given each token $\bm{x}$, we randomly sample a positive sample $\bm{c}$ from nearby tokens in the same context (sequence) within a window span where $W$ is the window size.

\textit{Negative pairs}: Given each token $\bm{x}$, we randomly sample $K$ tokens from other sequences in this batch as negative samples $\bm{c}_k^-$.

To sum up, the \textbf{T}oken-alignment \textbf{C}ontrastive (\textbf{TC}) loss is applied to every token in a batch as:
\begin{equation}
\small
    \label{tcloss1}
    \mathcal{L}_{\text{\objective}}
    =\frac1{N}\sum_{i=1}^{N}\frac1{|\bm{s}_i|}\sum_{j=1}^{|\bm{s}_i|} \mathcal{L}(\bm{g}_j^i, \bm{g}_{j_c}^i)
\end{equation}
where $N$ is the number of sequences of this batch; $\bm{s}_i$ is the $i$-th sequence; $j$ and $jc$ are tokens in $\bm{s}_i$ where $jc \neq j$; $\bm{g}^i$ is the global semantics hidden in contextualized representation of token $\bm{s}_i$. $\bm{g}_j^i$ and $\bm{g}_{j_c}^i$ are generated via:
\begin{align}
\small
\bm{g}_j^i &= \bm{h}_j^i - \bm{e}_j^i \\
\bm{g}_{j_c}^i &= \bm{h}_{j_c}^i - \bm{e}_{j_c}^i
\end{align}
where $\bm{h}_j^i$ and $\bm{e}_j^i $ are the contextualized representation and static embedding of the anchor token, respectively.  $\bm{h}_{j_c}^i$ and $\bm{e}_{j_c}^i$ are the contextualized representation and static embedding of the sampled positive token in the same context.

 \subsection{Training Objective}\label{subsec:tcmlm}
\label{subsec:semgrav:2}

As described before, the token-alignment contrastive loss $\mathcal{L}_{\text{\objective}}$ is designed to model global dependencies while MLM is able to capture local dependencies. Therefore, we can better model contextualized representations by combining the token-alignment contrastive loss $\mathcal{L}_{\text{\objective}}$ and the MLM loss to get our overall objective  $\mathcal{L}_{\text{\method}}$: 
\begin{equation}
\mathcal{L}_{\text{\method}}=\mathcal{L}_{\text{\objective}} + \mathcal{L}_{\text{MLM}}
\end{equation}
We implement it in a multi-task learning manner where all objectives are calculated within one forward propagation, which only introduces negligible extra computations.

\section{Experiments}
\label{sec:exps}
\subsection{Experimental Settings}

\paragraph{Training} 

Following BERT~\citep{bert}, we select the BooksCorpus (800M words after WordPiece tokenization)~\citep{zhu2015aligning} and English Wikipedia (4B words) as pre-training corpus.
We pre-train two variants of BERT models: BERT-small  and BERT-base. All models are equipped with the vocabulary of size 30,522, trained with 15\% masked positions for MLM. The maximum sequence length is 256 and  batch size is 1,280. We adopt optimizer AdamW~\citep{DBLP:conf/iclr/LoshchilovH19} with learning rate 1e-4. All models are trained until convergence. To be specific, the small model is trained up to 250k
steps with a warm-up of 2.5k steps. The base model is trained up to 500k steps with a warm-up of 10k steps. 
For \method, we set the positive sample window size $W$ to 5, the negative sample number $K$ to 50, and the temperature parameter $\tau$ to 0.07 after a slight grid-search via preliminary experiments. More pre-training details can be found in Appendix A.

During fine-tuning models, we conduct a grid search over batch sizes of \{16, 32, 64, 128\}, learning rates of \{1e-5, 2e-5, 3e-5, 5e-5\}, and training epochs of \{4, 6\} with an Adam optimizer~\citep{DBLP:journals/corr/KingmaB14}. We use the open-source packages for implementation, including HuggingFace Datasets\footnote{\scriptsize\url{https://github.com/huggingface/datasets}} and Transformers\footnote{\scriptsize\url{https://github.com/huggingface/transformers}}. All the experiments are conducted on 16 GPU chips (32 GB V100).

\begin{table*}[t]
\centering
\small

\begin{tabular}{clccccccccc}
\toprule
& Approach & MNLI(m/mm) & QQP & QNLI & SST-2 & CoLA & STS-B & MRPC & RTE & \textbf{Avg.} \\
\multirow{3}*{Validation Set} & MLM-${250k}$ & 76.9 / 77.4 & 85.7 & 86.2 & \textbf{89.0} & 28.8 & 85.6 & 85.9 & \textbf{59.6} & 75.0 \\
& \method-${50k}$ & 76.7 / 76.8 & 85.2 & 85.0 & 87.5 & 31.3 & 85.6 & 87.1 & 59.1 & 74.9 \\
& \method-${250k}$ & \textbf{77.9} / \textbf{78.4} & \textbf{86.1} & \textbf{86.5} & 88.9 & \textbf{34.2} & \textbf{86.1} & \textbf{88.1} & 59.5 & \textbf{76.2} \\
\midrule
\multirow{2}*{Test Set} & MLM-${250k}$ & 77.5~/~76.5 & \textbf{68.2} & 85.6 & 89.3 & 27.9 & 76.9 & 82.6 & \textbf{60.6} & 71.7 \\
& \method-${250k}$ & \textbf{78.0}~/~\textbf{76.9} & 67.6 & \textbf{86.3} & \textbf{89.5} & \textbf{31.2} & \textbf{77.8} & \textbf{84.4} & 58.4 & \textbf{72.2} \\
\bottomrule
\end{tabular}

\caption{\label{smallmodel}
GLUE results on BERT-small. For validation results, we run 4 experiments with different seeds for each task and report the average score. For test results, we report the test scores of the checkpoint performing best on validation sets. \method outperforms MLM with 1.2 point improvement and 5$\times$ speedup on validations sets. On test sets, \method also obtains better results on 6 out of 8 tasks.
}

\end{table*}

\paragraph{Evaluation}

We evaluate methods on the GLUE benchmark~\cite{glue}. Specifically, we test on Microsoft Research Paraphrase Matching (MRPC)~\citep{mrpc}, Quora Question Pairs (QQP)\footnote{\scriptsize\url{https://www.quora.com/q/quoradata/First-Quora-Dataset-Release-Question-Pairs}} and STS-B~\citep{senteval} for Paraphrase Similarity Matching; Stanford Sentiment Treebank (SST-2)~\citep{sst} for Sentiment Classification; Multi-Genre Natural Language Inference Matched (MNLI-m), Multi-Genre Natural Language Inference Mismatched (MNLI-mm)~\citep{mnli}, Question Natural Language Inference (QNLI)~\citep{qnli} and Recognizing Textual Entailment (RTE)~\citep{glue} for the Natural Language Inference (NLI) task; The Corpus of Linguistic Acceptability (CoLA)~\citep{cola} for Linguistic Acceptability. 

Following~\citet{bert}, we exclude WNLI~\citep{wnli}. We report F1 scores for QQP and MRPC, Spearman correlations for STS-B, and accuracy scores for the other tasks. For evaluation results on validation sets, we report the average score of 4 fine-tunings with different random seeds. For results on test sets, we select the best model on the validation set to evaluate.

\paragraph{Baselines}\label{baseline}
We mainly compare \method with MLM on BERT-small and BERT-base models. 
In addition, we also compare \method with related contrastive methods: a  sentence-level contrastive method BERT-NCE and a span-based contrastive learning method INFOWORD, both from ~\citet{Kong2020A}. We directly compare \method with the results reported in their paper. 

\subsection{Results on BERT-Small}

\begin{figure}
\centering
\small

\includegraphics[width=0.9\linewidth]{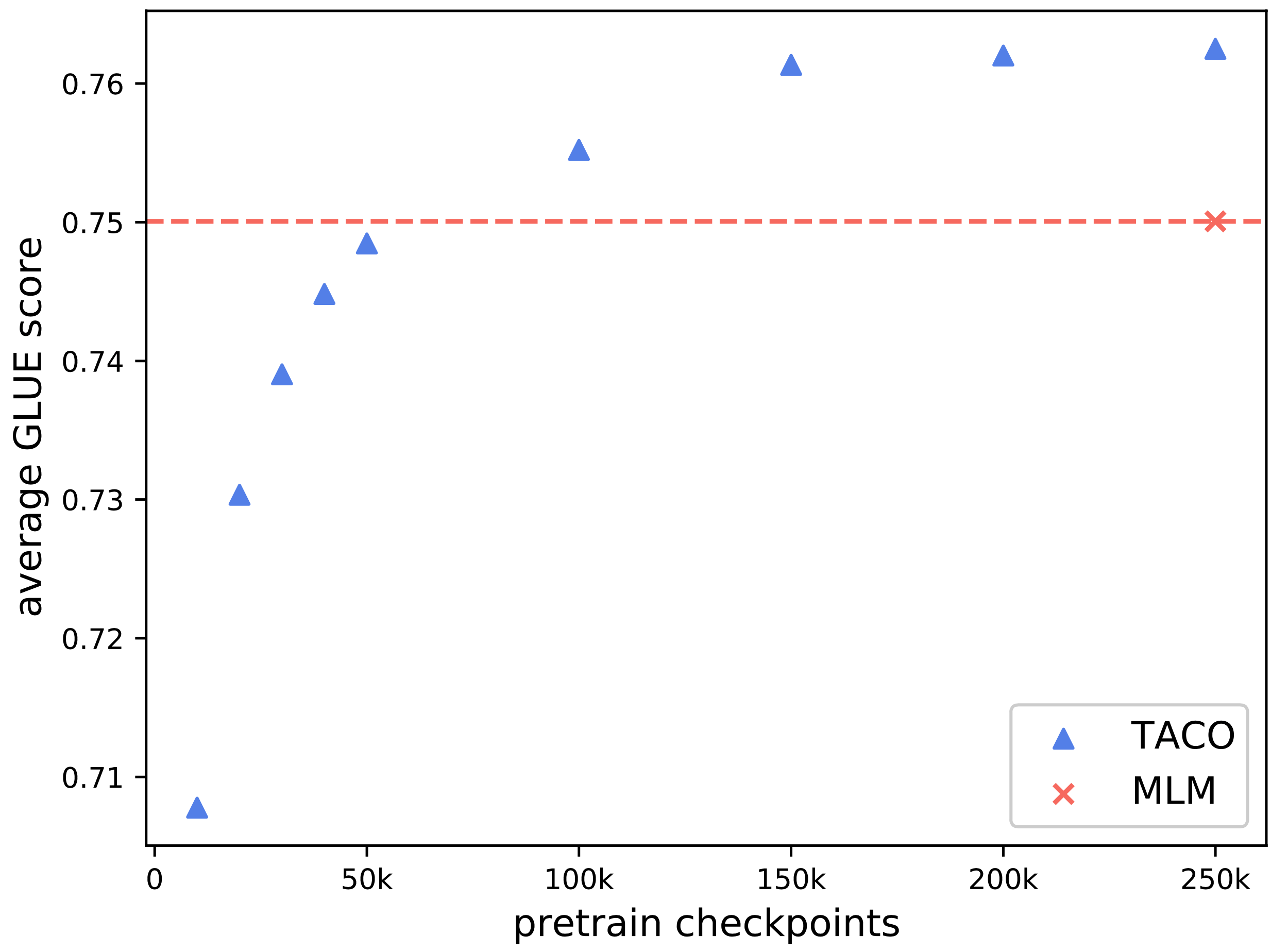}

\caption{Average GLUE score during pre-training. All results are reported on validation sets with BERT-small. \method achieves better results and 5$\times$ speedup than MLM.
}

\label{fig:smallmodel}
\end{figure}

Table \ref{smallmodel} and Figure \ref{fig:smallmodel} show the results of \method on BERT-small. As we can see, compared with MLM with 250k training steps ( convergence steps), \method achieves comparable performance with only 1/5 computation budget. By modeling global dependencies, \method can significantly improve the efficiency of contextualized representation learning. In addition, when pre-trained with the same steps, \method outperforms MLM with 1.2 average score improvement on the validation set. 

In addition to convergence, we also compare \method and MLM on fewer training data. The results are shown in Table~\ref{lessdata}. We sample 4 tasks with the largest amount of training data for evaluation. As we can see, TACO trained on 25\% data can achieve competitive results with MLM trained on full data. These results also verify the data efficiency of our method, \method.

\begin{table}[t]
\centering
\small

\begin{tabular}{lccccc}
\toprule
Approach & MNLI & QQP & QNLI & SST-2 & \textbf{Avg.} \\
\midrule
MLM-${25\%}$ & 77.8 & 85.7 & 85.8 & 87.2 & 84.1 \\
MLM-${100\%}$ & 76.9 & 85.7 & 86.2 & \textbf{89.0} & 84.5 \\
\method-${25\%}$ & 77.8 & 85.7 & 86.1 & 88.4 & 84.5 \\
\method-${100\%}$ & \textbf{77.9} & \textbf{86.1} & \textbf{86.5} & 88.9 & \textbf{84.9} \\
\bottomrule
\end{tabular}
\caption{\label{lessdata}
TACO pre-trained on a quarter of data achieves competitive downstream results with MLM pre-trained on full data.  All results are reported on GLUE validation sets with BERT-small. Here we sample 4 tasks with the largest amount of training data. 
}

\end{table}

\subsection{Results on BERT-Base}

We also compare \method with MLM on base-sized models, which are the most commonly used models according to the download data from Huggingface\footnote{\scriptsize\url{ https://huggingface.co/models}}~\citep{DBLP:conf/emnlp/WolfDSCDMCRLFDS20}. First, from Table \ref{basemodel}, we can see that \method consistently outperforms MLM under all pre-training computation budgets. Notably, \method-250$k$ achieves comparable performance with MLM-500$k$, which saves 2x computations. Similar results are observed on \method-100$k$ and BERT-250$k$. These results demonstrate that \method can achieve better acceleration over MLM. It is also a significant improvement compared to previous methods~\citep{DBLP:conf/icml/GongHLQWL19} focusing on accelerating BERT but only with slight speedups. In addition, as shown in Table \ref{basemodel-test}, \method achieves competitive results compared to BERT-NCE and INFOWORD, two similar contrastive methods.

\begin{table*}[t]
\centering
\small

\begin{tabular}{lccccccccc}
\toprule
Approach & MNLI & QQP & QNLI & SST-2 & CoLA & STS-B & MRPC & RTE & \textbf{Avg.}\\
\midrule
MLM-${100k}$ & 80.7 & 86.4 & 89.3 & 90.5 & 47.4 & 86.0 & 85.0 & 56.6 & 77.7 \\
MLM-${250k}$ & 83.0 & 87.4 & 90.4 & 91.8 & 48.6 & 87.1 & 87.5 & 57.8 & 79.2 \\
MLM-${500k}$ & 84.2 & 87.9 & \textbf{91.1} & 92.1 & 51.1 & 87.9 & 89.8 & 63.4 & 80.9 \\
\midrule
\method-${100k}$ & 81.5 & 87.4 & 89.4 & 90.3 & 46.4 & 87.2 & 87.8 & 62.8 & 79.1 \\
\method-${250k}$ & 83.8 & 87.9 & 90.2 & 91.4 & 50.7 & 87.9 & 89.3 & 63.5 & 80.6 \\
\method-${500k}$ & \textbf{84.6} & \textbf{88.1} & 90.8 & \textbf{92.3} & \textbf{53.4} & \textbf{88.5} & \textbf{90.7} & \textbf{66.3} & \textbf{81.8} \\
\bottomrule
\end{tabular}

\caption{
GLUE results on BERT-base. All results are reported on validation sets. We run 6 experiments with different hyper-parameter combinations (including random seeds) for each task and report the average score. 
The MNLI-matched score is reported here. TACO outperforms MLM with 0.9 point improvement and 2$\times$ speedup. 
}

\label{basemodel}
\end{table*}

\begin{table*}[h]
\centering
\small

\begin{tabular}{lccccc}

\toprule
Approach & MNLI(m/mm) & QQP & QNLI & SST-2 & \textbf{Avg.} \\
\midrule
BERT-NCE & 83.2~/~83.0 & 70.5 & 90.9 & 93.0 & 84.1 \\
INFOWORD & 83.7~/~82.4 & 71.0 & 91.4 & 92.5 & 84.2 \\
\method & \textbf{84.5}~/~\textbf{83.5} & \textbf{71.7} & \textbf{91.6} & \textbf{93.2} & \textbf{84.9} \\
\bottomrule

\end{tabular}

\caption{
\method achieves the best among contrastive-based methods. All results are reported on GLUE test sets with BERT-base. For each task, we report test results of the checkpoint performing best on validation sets. 
}

\label{basemodel-test}
\end{table*}

\section{Discussion}
\label{sec:ana}
\subsection{\method and MLM}

\begin{figure*}[t]
\centering
\small

\includegraphics[width=0.45\linewidth]{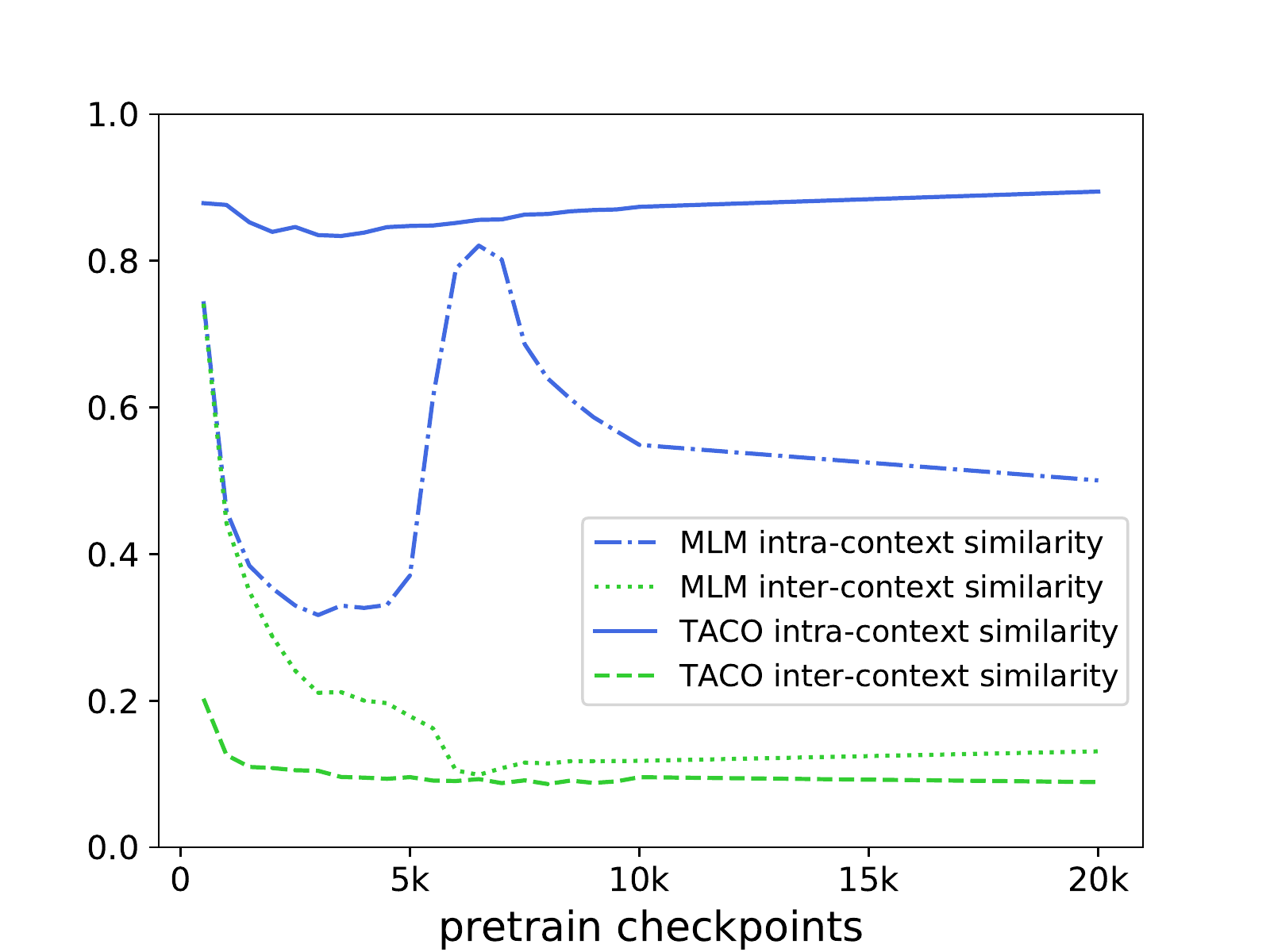}
\includegraphics[width=0.45\linewidth]{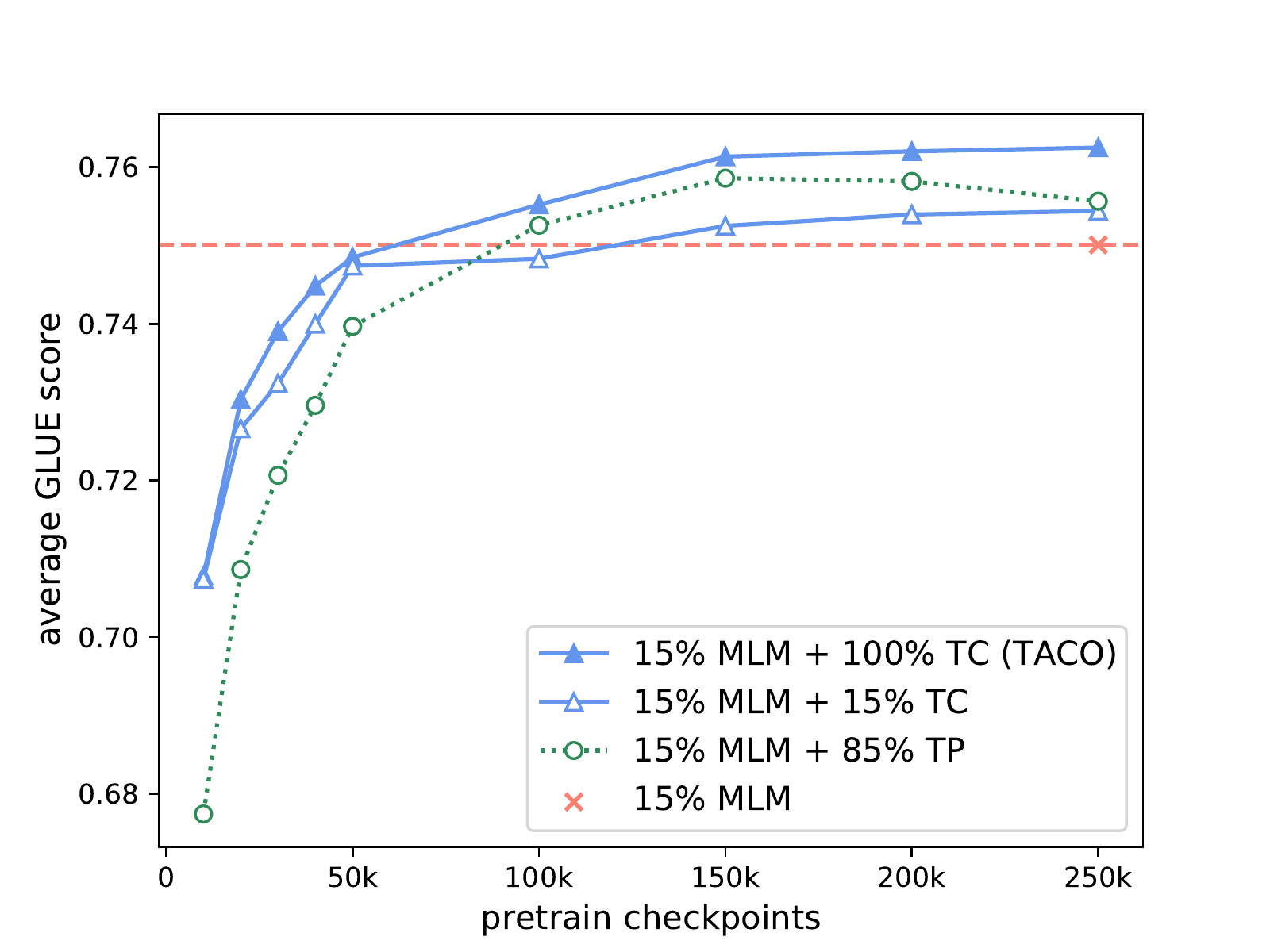}

\caption{The left figure (a) shows the intra-context similarity and inter-context similarity change during pre-training.  The right figure (b) shows two ablations of \method: a concentrated \method (15\% MLM + 15\% TC), where the MLM loss and the TC loss are both built on the same 15\% masked positions, and an extended MLM (15\% MLM + 85\% TP), which masks 15\% positions but predict original tokens on all positions. 
}

\label{fig:analysis}
\end{figure*}

To better understand how \method works, we conduct a quantitative comparison on the learning dynamic for BERT and \method. Similar to Section \ref{sec:analysis}, we plot the Cosine similarity among contextualized representations of tokens in the same context (intra-context) and different contexts (inter-context) in Figure \ref{fig:analysis}. We find that the learning dynamic of \method significantly differs from that of MLM. Specifically, for \method, the intra-context representation similarity remains high and the gap between intra-context similarity and inter-context similarity remains large at the later stage of training. This confirms that \method can better fulfill global semantics, which may contribute to the superior downstream performance.

\subsection{Ablation Study}

\method is implemented as a token-level contrastive (TC) loss along with the MLM loss. Therefore, the improvement of \method might come from two aspects, including 1) denser supervision signals from the all-token objective and 2) the benefits of the contrastive loss to strengthen global dependencies. It is helpful to figure out which factor is more important. To this end, we design two variants for ablation. One is a \textit{concentrated} \method, where the contrastive loss is built on the 15\% masked positions only, keeping the same density of supervision signal with MLM. The other is an \textit{extended} MLM, where not only 15\% masked positions are asked to predict the original token, so do the rest 85\% unmasked positions. The extended MLM has the same dense supervision with \method but loses the benefits of modeling the global dependencies. The results on small models are shown in Figure \ref{fig:analysis}.

As we can see, the performance of \method decreases if we sample a part of token positions to implement TC objectives. It shows that more supervision signals benefit the final performance of \method. However, simply adding more supervision signals by predicting unmasked tokens does not help MLM too much. Even equipped with the extra 85\% token prediction (TP) loss, MLM+TP does not show significant improvements and it is noticeable that the performance of  MLM+TP starts to drop after 150k steps. This further confirms the effectiveness of TC loss by strengthening global dependencies. 

\section{Related Work}
\label{sec:related}
\subsection{Language Representation Learning}

Classic language representation learning methods~\citep{DBLP:conf/nips/MikolovSCCD13,DBLP:conf/emnlp/PenningtonSM14} aims to learn context-independent representation of words, i.e., word embeddings. They generally follow the distributional hypothesis~\citep{harris1954distributional}. 
Recently, the pre-training then fine-tuning paradigm has become a common practice in NLP because of the success of pre-trained language models like BERT~\citep{bert}. Context-dependent (or contextualized) representations are the basic characteristic of these methods. Many existing contextualized models are based on the masked language modeling objective, which randomly masks a portion of tokens in a text sequence and trains the model to recover the masked tokens. Many previous studies prove that pre-training with the MLM objective helps the models learn syntactical and semantic
knowledge~\citep{DBLP:journals/corr/abs-1906-04341}. There have been numerous extensions to MLM. For example, XLNet~\citep{DBLP:conf/nips/YangDYCSL19} introduced the permutated language modeling objective, which predicts the words one by one in a permutated order. BART~\cite{DBLP:conf/acl/LewisLGGMLSZ20} and T5~\cite{DBLP:journals/jmlr/RaffelSRLNMZLL20} investigated several denoising objectives and pre-trained an encoder-decoder architecture with the mask span infilling objective. In this work, we focus on the key MLM objective and aim to explore how MLM objective helps learn contextualized representation. 

\subsection{Contrastive-based SSL}

Apart from denoising-based objectives, contrastive learning is another promising way to obtain self-supervision. In contrastive-based self-supervised learning, the models are asked to distinguish the positive samples from the negative ones for a given anchor. Contrastive-based SSL method was first introduced in NLP for efficient learning of word representations by negative sampling, i.e., SGNS (Word2Vec~\citep{DBLP:conf/nips/MikolovSCCD13}). Later, similar ideas were brought into CV field for learning image representation and got prevalent, such as MoCo~\cite{DBLP:conf/cvpr/He0WXG20}, SimCLR~\cite{DBLP:conf/icml/ChenK0H20}, BYOL~\cite{DBLP:conf/nips/CaronMMGBJ20}, etc.

In the recent two years, there have been many studies targeting at reviving contrastive learning for contextual representation learning in NLP. For instance, CERT~\citep{fang2020cert} utilized back-translation to generate positive pairs. CAPT~\citep{luo2020capt} applied masks to the original sentence and considered the masked sentence and its original version as the positive pair. DeCLUTR~\citep{giorgi2020declutr} samples nearby even overlapping spans as positive pairs. INFOWORD~\citep{Kong2020A} treated two complementary parts of a sentence as the positive pair. However, the aforementioned methods mainly focus on sentence-level or span-level contrast and may not provide dense self-supervision to improve efficiency. Unlike these approaches, \method regards the global semantics hidden in contextualized token representations as the positive pair. The token-level contrastive loss can be built on all input tokens, which provides a dense self-supervised signal. 

Another related work is ELECTRA~\citep{DBLP:conf/iclr/ClarkLLM20}. ELECTRA samples machine-generated tokens from a separate generator and trains the main model to discriminate between machine-generated tokens and original tokens. ELECTRA implicitly treats the fake tokens as negative samples of the context, and the unchanged tokens as positive samples. Unlike this method, \method does not require architectural modifications and can serve as a plug-and-play auxiliary objective, largely improving pre-training efficiency.  



\section{Conclusion}
\label{sec:conclusion}
In this paper, we propose a simple yet effective objective to learn contextualized representation. 
Taking MLM as an example, we investigate whether and how current language model pre-training objectives learn contextualized representation. 
We find that the MLM objective mainly focuses on local anchors to align contextualized representations, which harms global dependencies modeling due to an ``embedding bias'' problem. Motivated by these problems, we propose TACO to directly model global semantics. It can be easily combined with existing LM objectives. By combining local and global anchors, \method achieves up to 5$\times$ speedups and up to 1.2 improvements on GLUE score. This demonstrates the potential of \method to serve as a plug-and-play approach to improve contextualized representation learning.

\section*{Acknowledgement}
\label{sec:acknowledgement}
We thank the anonymous reviewers
for their helpful feedback. We also thank the colleagues from ByteDance AI Lab for their suggestions on our experiment designing and paper writing.

\bibliography{paper}
\bibliographystyle{acl_natbib}

\clearpage
\appendix
\section{Experiment Details}
\label{sec:appendix}

\begin{table*}[hb]
\centering
\small

\begin{tabular}{llll}
\toprule
Pre-training & Hyper-parameters & Small & Base \\
\midrule

\multirowcell{25}{Parameters Shared by\\All Approaches} & Number of Layers & 4 & 12 \\
& Hidden Size & 512 & 768 \\
& Hidden Layer Activation Function & gelu & gelu \\
& FFN Inner Hidden Size & 2,048 & 3,072 \\
& Attention Heads & 8 & 12 \\
& Attention Head Size & 64 & 64 \\
& Embedding Size & 512 & 768 \\
& Vocab Size & 30,522 & 30,522 \\
& Max Position Embeddings & 512 & 512 \\
& Max Sequence Length & 256 & 256 \\
& Attention Dropout & 0.1 & 0.1 \\
& Dropout & 0.1 & 0.1 \\
& Initializer Range & 0.02 & 0.02 \\
& Learning Rate Decay & Linear & Linear \\
& Learning Rate & 1e-4 & 1e-4 \\
& Max Gradient Norm & 1.0 & 1.0 \\
& Adam $\epsilon$ & 1e-8 & 1e-8 \\
& Adam $\beta_1$ & 0.9 & 0.9 \\
& Adam $\beta_2$ & 0.999 & 0.999 \\
& Weight Decay & 0.01 & 0.01 \\
& Batch Size & 1,280 & 1,280 \\
& Train Steps & 250k & 500k \\
& Warm-up Steps & 2,500 & 10,000 \\
& FP16 & True & True \\
& Mask Percentage & 15 & 15 \\
\midrule
\multirowcell{3}{\method\\Only} & Negative Sample Size $K$ & 50 & 50 \\
& Positive Sample Window Size $W$ & 5 & 5 \\
& Temperature Parameter $\tau$ & 0.07 & 0.07 \\

\bottomrule
\end{tabular}

\caption{Hyper-parameters during pre-training.}

\label{appendix:pretrain}
\end{table*}

\subsection{Pre-training Hyper-parameters}
All pre-training approaches involved in experiments use the same pre-training hyper-parameters but do not include BERT-NCE and INFOWORD. Results of BERT-NCE and INFOWORD are directly cited from the original paper~\citep{Kong2020A}. 
Following \citet{roberta}, we do not use the next sentence prediction (NSP) objective and use dynamic masking for MLM with a 15\% mask ratio, where the masked positions are decided on the fly.

\method introduces three extra hyper-parameters, including negative sample size $K$, positive sample window size $W$ and temperature $\tau$. We set the temperature  $\tau$ as a small value, 0.07, following \citet{fang2020cert}. By searching for the best $K$ out of \{10, 50\} and $W$ out of \{3, 5, 10, 50\} on the small \method model, we found that \method with $K$=50 and $W$=5 performs best, so we also apply these hyper-parameter choices for base-sized \method. The full set of pre-training hyper-parameters are listed in Table \ref{appendix:pretrain}. Actually, TACO outperforms MLM under most cases in our preliminary experiments. However, we still also find some extreme cases which might harm the effectiveness of TACO. If the size of negative samples $K$ is too small, e.g., smaller than 10, the performance of TACO degenerates nearly to the level of BERT baseline. Similar conclusions are also mentioned in related works \cite{DBLP:conf/cvpr/He0WXG20, DBLP:conf/icml/ChenK0H20}. Also, if the positive window size $W$ is too large, e.g., bigger than 50, the performance of TACO degrades, too. We suspect the over-large positive window brings more false-positive samples, which makes the sequence meaning ambiguous, thus harms the performance.

\subsection{Fine-tuning Details}

\begin{table*}[ht]
\centering
\small

\begin{tabular}{lll}
\toprule
Fine-tuning & Hyper-parameters & Small/Base\\
\midrule

\multirowcell{11}{Parameters Shared by \\All Models} & Max Sequence Length & 128\\
& Attention Dropout & 0.1\\
& Dropout & 0.1\\
& Initializer Range & 0.02\\
& Learning Rate Decay & Linear\\
& Max Gradient Norm & 1.0\\
& Adam $\epsilon$ & 1e-8\\
& Adam $\beta_1$ & 0.9\\
& Adam $\beta_2$ & 0.999\\
& Weight Decay & 0.0\\
& FP16 & False\\

\bottomrule
\end{tabular}

\caption{Hyper-parameters during fine-tuning.}

\label{appendix:finetune_common}
\end{table*}

For small-sized models, we fine-tune all saved checkpoints (5k, 10k, 20k, 30k, 40k, 50k, 100k, 150k, 200k, 250k-step) of different pre-trained models (\method and its ablations) with the same hyper-parameters on each task. Considering the large amount of pre-training checkpoints, we just adopt the default fine-tuning hyper-parameters and repeat fine-tuning 4 times with different random seeds. Then the best performed fine-tuned models on validation sets are used for testing.
This setting helps make a fair comparison among models and avoids a large amount of grid-search runs. The task-specific hyper-parameters for small-sized models are listed in Table \ref{appendix:finetune_small_specific}. The general fine-tuning hyper-parameters are listed in Table \ref{appendix:finetune_common}.

\begin{table*}[t]
\centering
\small

\begin{tabular}{lcccc}
\toprule
Task & Learning Rate & Batch Size & Train Epochs & Warm-up Steps \\
\midrule
MNLI & 5e-5 & 64 & 6 & 2,000 \\
QQP & 5e-5 & 64 & 6 & 2,000 \\
QNLI & 5e-5 & 64 & 4 & 200 \\
SST-2 & 5e-5 & 64 & 4 & 200 \\
CoLA & 5e-5 & 32 & 4 & 100 \\
STS-B & 5e-5 & 32 & 4 & 100 \\
MRPC & 5e-5 & 32 & 4 & 100 \\
RTE & 5e-5 & 32 & 4 & 100 \\

\bottomrule
\end{tabular}

\caption{Task-specific hyper-parameters for small models during fine-tuning.}

\label{appendix:finetune_small_specific}
\end{table*}

For base-sized models, we save checkpoints at 100k, 250k, and 500k steps, respectively. 
During fine-tuning, we also conduct multiple fine-tuning runs with different task-specific hyper-parameter combinations as shown in Table \ref{appendix:finetune_base_specific}. 
Concretely, we randomly sample 6 different hyper-parameter combinations and report the average score for validation results. Then we select the best-performing run of 500k-step checkpoints (converged) for testing.

\begin{table*}[t]
\centering
\small

\begin{tabular}{lcccc}
\toprule
Task & Learning Rate & Batch Size & Train Epochs & Warm-up Steps \\
\midrule
MNLI & \{1e-5, 2e-5, 3e-5, 5e-5\} & \{32, 64, 128\} & \{4, 6, 8\} & \{1000, 2000\} \\
QQP & \{1e-5, 2e-5, 3e-5, 5e-5\} & \{32, 64, 128\} & \{4, 6, 8\} & \{1000, 2000\} \\
QNLI & \{1e-5, 2e-5, 3e-5, 5e-5\} & \{32, 64\} & \{4, 6\} & \{100, 200, 1000\} \\
SST-2 & \{1e-5, 2e-5, 3e-5, 5e-5\} & \{16, 32, 64\} & \{4, 6\} & 200 \\
CoLA & \{1e-5, 2e-5, 3e-5, 5e-5\} & \{16, 32, 64\} & \{4, 6\} & 100 \\
STS-B & \{1e-5, 2e-5, 3e-5, 5e-5\} & \{16, 32, 64\} & \{4, 6\} & 100 \\
MRPC & \{1e-5, 2e-5, 3e-5, 5e-5\} & \{16, 32, 64\} & \{4, 6\} & 100 \\
RTE & \{1e-5, 2e-5, 3e-5, 5e-5\} & \{16, 32, 64\} & \{4, 6, 8\} & 100 \\

\bottomrule
\end{tabular}

\caption{
Task-specific hyper-parameters for base models during  fine-tuning.
}

\label{appendix:finetune_base_specific}
\end{table*}

\subsection{Statistic Details}

\begin{table*}[t]
\centering
\small
\begin{tabular}{lllllllllll}
\toprule
Measurement / Checkpoint & 1k & 2k & 3k & 5k & 7.5k & 10k & 20k & 50k & 100k & 250k\\
\midrule
L1 Distance & 0.977 & 0.925 & 0.880 & 0.833 & 0.769 & 0.779 & 0.774 & 0.797 & 0.820 & 0.838\\
L2 Distance & 0.978 & 0.927 & 0.884 & 0.838 & 0.778 & 0.789 & 0.783 & 0.803 & 0.826 & 0.843\\
L10 Distance & 0.981 & 0.928 & 0.890 & 0.854 & 0.802 & 0.811 & 0.805 & 0.822 & 0.844 & 0.860\\
Cosine Similarity & 1.093 & 1.314 & 1.548 & 1.890 & 3.197 & 3.533 & 3.591 & 3.482 & 3.325 & 3.174\\
Dot-production Similarity & 1.092 & 1.313 & 1.547 & 1.890 & 3.189 & 3.525 & 3.586 & 3.480 & 3.321 & 3.166\\
\bottomrule
\end{tabular}

\caption{
The ratio of intra-context measurement over inter-context measurement during pre-training. We list two distance measurements and three similarity measurements here. 
}

\label{appendix:statistics_other_measurements}
\end{table*}

\paragraph{Embedding Similarity} We calculate cosine similarity of 20 randomly sampled pairs of frequently co-occurrent words from the WordSim353 dataset~\citep{agirre2009study} labeled by human annotators to plot the average similarity curve in Figure \ref{fig:similarity}(b). Corresponding embeddings are obtained from the embedding layer of the BERT model and variant models mentioned in Section \ref{sec:analysis}.

\paragraph{Intra-/Inter-context Similarity} 
For every token $w_i$ in the corpus, we randomly sample a positive token $w_{j\ne i}$ within the same context (sentence) and another token $w_{k}$ from other sentences.

As mentioned in Section~\ref{sec:analysis}, we take BERT~\cite{bert} as our encoder to get contextualized representations through the last hidden states $\bm{h}$. We mainly adopt the cosine similarity as the measurement and calculate the average intra-context similarity (between $\bm{h}_i$ and $\bm{h}_j$) and the average inter-context similarity (between $\bm{h}_i$ and $\bm{h}_k$) over all tokens in the corpus. It is worth noticing that we do use any masks here when generating a token's contextualized representation for statistics. 

\paragraph{Other Measurements} We observe the same findings for MLM under other measurements, though the statistics before are mainly based on cosine similarities. We tried other similarities or distances, e.g., L1 distance, L2 distance and L10 distance, to evaluate the discrepancy between contextualized representations from the same context and different contexts. Specifically, we make intra-context and inter-context statistics under specific measurement at different pre-training checkpoints, then calculate the \text{ratio} of intra-context measurement over the inter-context one. Table \ref{appendix:statistics_other_measurements} shows the statistical results. As we can see, when the ratio of L1 distance decreases, the ratio of cosine similarity and the dot-production similarity increase, vice versa. 

\section{Extra Experiments}
\label{sec:appendix_extra}
\begin{table*}[b]
\centering
\small

\begin{tabular}{lccccccccc}
\toprule
Approach & MNLI & QQP & QNLI & SST-2 & CoLA & STS-B & MRPC & RTE & \textbf{Avg.}\\
\midrule
MLM-${250k}$ & 76.9 / 77.4 & 85.7 & 86.2 & \textbf{89.0} & 28.8 & 85.6 & 85.9 & \textbf{59.6} & 75.0 \\ 
 \method-${50k}$ & 76.7 / 76.8 & 85.2 & 85.0 & 87.5 & 31.3 & 85.6 & 87.1 & 59.1 & 74.9 \\
  \method-${50k}$  w/o shared embedding & 76.3 / 76.5 & 85.0 & 85.2 & 87.2 & 32.5 & 85.1 & 86.7 & 58.9 & 74.6 \\ 
 \method-${250k}$ & \textbf{77.9} / \textbf{78.4} & 86.1 & \textbf{86.5} & 88.9 & 34.2 & \textbf{86.1} & \textbf{88.1} & 59.5 & \textbf{76.2} \\
 \method-${250k}$  w/o shared embedding & 77.5 / 78.2 & \textbf{86.3} & 86.2 & 88.5 & \textbf{35.1} & 85.8 & 88.0 & 59.3 & 75.9 \\
\bottomrule
\end{tabular}

\caption{
Results on GLUE validation set with small-size models. For models without embedding sharing, we run 3 experiments with different random seeds for each task and report the average score.
}

\label{appendix:sharing_embedding}
\end{table*}

In the standard implementation of BERT, the parameters of input embeddings are shared with output embeddings. All experiments and analyses in this paper are based on this assumption. To further confirm the effectiveness of \method, we conduct the extra experiments without embedding sharing on BERT-small. The results are showed in Table \ref{appendix:sharing_embedding}. It is unexpected that the variants without embedding sharing perform worse compared their counterparts due to lack of regularization of weight sharing. From the results, we can see that the TACO without embedding sharing performs slightly worse than TACO with embedding sharing. However, compared to the MLM, it is still better than MLM than 0.9 average GLUE score when convergence. These results prove the effectiveness of TACO even when embeddings are not sharing.

\end{document}